\documentclass{IEEEcsmag}

\usepackage[colorlinks,urlcolor=blue,linkcolor=blue,citecolor=blue]{hyperref}
\expandafter\def\expandafter\UrlBreaks\expandafter{\UrlBreaks\do\/\do\*\do\-\do\~\do\'\do\"\do\-}
\usepackage{upmath,color}
\usepackage{cite}

\jname{IEEE Intelligent Systems}
\jtitle{IEEE Intelligent Systems}
\pubyear{2026}

\begin{document}

\sptitle{\footnotesize This article has been accepted for publication in IEEE Intelligent Systems. This is the author's version which has not been fully edited and content may change prior to final publication. Citation information: DOI 10.1109/MIS.2026.3660833}

\title{Multiple Choice Questions: Reasoning Makes Large Language Models (LLMs) More Self-Confident, \textcolor{white}{Especially} When They are Wrong}

\author{Tairan Fu}
\affil{Politecnico di Milano, Milan, Italy}

\author{Javier Conde}
\affil{Information Processing and Telecommunications Center (IPTC), Universidad Politécnica de Madrid, Madrid, 28040, Spain}

\author{Gonzalo Martínez}
\affil{Information Processing and Telecommunications Center (IPTC), Universidad Politécnica de Madrid, Madrid, 28040, Spain}

\author{María Grandury}
\affil{Information Processing and Telecommunications Center (IPTC), Universidad Politécnica de Madrid, Madrid, 28040, Spain}

\author{Pedro Reviriego}
\affil{Information Processing and Telecommunications Center (IPTC), Universidad Politécnica de Madrid, Madrid, 28040, Spain}

\markboth{%
\parbox{\textwidth}{\scriptsize
This article has been accepted for publication in IEEE Intelligent Systems.
This is the author's version which has not been fully edited \\ and content may change prior to final publication. Citation information: DOI 10.1109/MIS.2026.3660833
}}{%
\parbox{\textwidth}{\scriptsize
This article has been accepted for publication in IEEE Intelligent Systems.
This is the author's version which has not been fully edited \\ and content may change prior to final publication. Citation information: DOI 10.1109/MIS.2026.3660833
}}

\begin{abstract}\looseness-1Multiple Choice Question (MCQ) tests are among the most used methods for evaluating large language models (LLMs). Besides checking the correctness of the selected answer, evaluations often consider the model’s confidence through the probability assigned to its response. In this work, we investigate how LLM confidence is influenced by the answering approach when the model answers directly or reasons before responding. \textcolor{black}{Experiments on a general knowledge benchmark, covering 57 subjects and seven LLMs, show that models are systematically more confident when providing reasoning before answering, and that this confidence increase is larger when the selected answer is incorrect than when it is correct.} We hypothesize that the reasoning process alters token probabilities, as the final answer prediction depends jointly on the question and the model’s self-generated reasoning, leading to inflated confidence estimates. \textcolor{black}{Using standard calibration metrics such as Expected Calibration Error and Brier score, we further show that Chain-of-Thought (CoT) prompting degrades calibration by increasing the proportion of high-confidence wrong answers. These findings indicate that, in MCQ evaluation settings with CoT prompting, LLM-estimated probabilities should be used with caution as a basis for evaluation and metacognitive mechanisms.}
\end{abstract}

\maketitle


\begingroup
\renewcommand\thefootnote{}
\footnotetext{This work is licensed under a Creative Commons Attribution 4.0 License. For more information, see https://creativecommons.org/licenses/by/4.0/}
\addtocounter{footnote}{-1}
\endgroup

\chapteri{T}\textcolor{black}{he evaluation of Large Language Models (LLMs) is challenging, as in most cases their answers are in natural language and they have to be evaluated on a large number of topics and tasks \cite{LLM_evaluation_survey}. Even when evaluation uses structured or constrained outputs (for example, JSON, or a structured answer tag as in this work), the question, reasoning and evidence are expressed in natural language, complicating systematic, scalable assessment.} A potential approach is human evaluation, so people evaluate LLM responses. However, this does not scale to tens of thousands of questions for each model, with new models appearing every day. To address this issue, initiatives such as Chatbot Arena \cite{chatbot_arena} resort to the community to assess human preferences. However, questions, answers, and participants are not controlled, so the results provide a comparative ranking of models, but not a detailed analysis of their specific capabilities. A more scalable alternative would be to use an LLM to evaluate other LLMs \cite{llmjudging}. This method has limitations, as the LLM that is judging may have biases towards its own content or toward long responses, and someone has to evaluate this LLM. Today, the most widely used method to evaluate LLMs is to run different benchmarks that are mostly made up of multiple-choice questions. This enables automation of the process and evaluation of specific tasks, for example, mathematics, reasoning, or knowledge of many different topics \cite{MMLU}. 

The results of a Multiple Choice Question (MCQ) test are typically measured in terms of the percentage of correct answers using a given number of examples in the prompt to help the model \cite{LLM_evaluation_survey}. This accuracy metric does not provide any insight into the confidence of the LLM in its responses, which is an important feature of the LLM \cite{confidence_survey}. However, as in order to select each new token, an LLM computes estimates of the probability that each token in its dictionary is the next token, these probabilities can be used to develop confidence estimates \cite{logit_uncertainty_calibration}. For example, if there are four possible options to answer a question, \textit{A,B,C,D}, and the LLM estimated probabilities for each of them are \textit{0.5,0.3,0.2,0.1}, the model does not have much confidence in its response. However, these are just the estimated probabilities of the LLM and may not correlate with the correctness of each option, as LLMs have limited metacognition capabilities \cite{steyvers2025metacognition}.

The LLM responses depend not only on the question but also on the text produced by the LLM before selecting an option. In fact, it is well known that in many cases LLMs have better results when they are asked to think and decompose the problem into several steps, a technique known as Chain of Thoughts (CoT) \cite{Chain-of-Thought}. This can be done in MCQ tests by asking the LLM to first provide the reasoning and then select an option on the prompt. An interesting question is whether this reasoning has any impact on the confidence of the LLM in its choice. In more detail, are the LLM estimated probabilities for its selected option different when the model reasons from those of when the model answers directly? If there is a difference, does it apply to choices that are correct, wrong, or to both? Is the use of logprobs, therefore a recommended measure to evaluate the confidence level of LLMs? In which cases is it recommended? Are logprobs a useful tool for the study of LLM metacognition when answering MCQs?

In this paper, \textcolor{black}{we focus on the metacognitive behavior of LLMs when they solve multiple-choice questions presented in natural language. In particular, we study how} the self-confidence of LLMs in their MCQ responses varies when the models answer directly or when they first provide the reasoning and then the selected answer. The main findings of our evaluation show that 

\begin{enumerate}
    \item Models tend to be more confident when they reason before answering.
    \item This increase in self-confidence occurs both when the model response is correct and when it is incorrect, \textcolor{black}{and is systematically larger when the response is incorrect.}
    \item \textcolor{black}{Chain-of-Thought prompting worsens standard calibration metrics for most models by increasing the proportion of high-confidence wrong answers.}    
    \item The increase occurs for practically all categories/topics of the questions, but it is larger for topics that require reasoning.
    \item All tested models experience similar trends across all dimensions.
\end{enumerate}

In the paper, these effects are discussed and linked to the operation of LLMs and also to human cognition to try to understand their causes and implications for the study of metacognition in LLMs.


\section{CONFIDENCE AND REASONING}
\label{sec:Related work}
\subsection{Confidence estimation}

With the widespread adoption of LLMs across various tasks, reliably assessing the confidence of their outputs has become a critical challenge. This issue is closely tied to the practical usability of such models. Several methods have been proposed for confidence estimation, including: (1) approaches based on token logprobs; (2) prompting-based or fine-tuning-based techniques that elicit self-evaluation from the model \cite{selfevaluation}; and (3) strategies involving multiple sampling or collaboration among different models \cite{sampleconfidence}.

Among these, methods based on token logprobs are particularly popular due to their simplicity, efficiency, and minimal computational overhead \cite{logitconfidence1}. By analyzing the logpros distribution produced during token generation, these approaches quantify the model's ``hesitation'' in forming a response, thereby reflecting its confidence in the final output. A recent work \cite{disentangleuncertainty} further advances this direction by disentangling uncertainty from the logpros distribution into aleatoric and epistemic components, offering a more systematic and interpretable framework for confidence evaluation. Additionally, some studies have pointed out that, due to the autoregressive nature of LLMs, their responses consist of many sequential tokens, yet a small subset of critical tokens can determine the overall direction of the output \cite{Bigelow2024ForkingPI}. These observations further underscore the need for more fine-grained research into confidence estimation for LLMs and its implications for LLM metacognition.

\subsection{Reasoning: Chain of thought}

CoT is a prompting strategy that guides language models to generate intermediate reasoning steps before arriving at a final answer, thereby enhancing their reasoning capabilities and interpretability. Since its introduction \cite{Chain-of-Thought}, CoT has been widely applied to various tasks such as mathematical reasoning and symbolic reasoning \cite{zero-shot} and has significantly improved the performance of LLMs on complex and multi-step problems.

However, alongside these improvements, CoT also introduces certain challenges. Studies have shown that it can lead to overthinking, where models expend unnecessary computational resources on simple problems \cite{Chen2024DoNT} and may even follow incorrect reasoning paths that result in wrong answers \cite{Zhu2025UncertaintyGuidedCF}. While considerable research has focused on the performance implications of CoT, its impact on confidence estimation remains underexplored.

This gap is particularly concerning. A logically coherent reasoning process can already be highly persuasive to human users. If a model not only produces an incorrect conclusion but also assigns it high confidence, the erroneous output is more likely to be accepted without scrutiny.

\section{ESTIMATING CONFIDENCE}
\label{sec:methodology}

This section describes the methodology used in the evaluation including the experimental procedure, the LLMs evaluated, and the benchmarks considered.

\subsection{Procedure}

In our evaluation, we consider two different prompts when asking the question to the model. In the first, the model is asked to answer directly:

\textit{``Please respond with only the letter of the solution, in the format \{`sol': `solution'\}. Do not respond with any other information. Here is an example: 
\\
Input: A car travels 60 kilometers per hour for 2 hours and then 80 kilometers per hour for 3 hours. What is the average speed of the car for the entire trip? a) 70 km/h, b) 72 km/h, c) 75 km/h, d) 74 km/h
\\
Output: \{`sol': `b'\}''}

In the second prompt, as per CoT, the model is asked to provide step-by-step reasoning before selecting an option:

\textit{``Please think step by step before answering, considering at least three steps. Once you have the solution, end the response only with the letter of the solution, in the format \{`sol': `solution'\}. Here is an example:
\\
Input: A car travels 60 kilometers per hour for 2 hours and then 80 kilometers per hour for 3 hours. What is the average speed of the car for the entire trip? a) 70 km/h, b) 72 km/h, c) 75 km/h, d) 74 km/h
\\
Output: First, I need to calculate the total distance traveled. For the first part of the trip, the car travels at 60 km/h for 2 hours, so the distance is 60 * 2 = 120 kilometers. Next, for the second part of the trip, the car travels at 80 km/h for 3 hours, so the distance is 80 * 3 = 240 kilometers. The total distance traveled is 120 + 240 = 360 kilometers. Now, I need to calculate the total time spent. The total time is 2 + 3 = 5 hours. To find the average speed, I divide the total distance by the total time: 360 kilometers ÷ 5 hours = 72 km/h. Therefore, the correct answer is \{`sol': `b'\}''.}

\textcolor{black}{The format \{`sol': `solution'\} is used only for evaluation convenience and does not change the fact that the question and reasoning process are in natural language. This format is used to facilitate the parsing of the responses} to extract the option selected by the model and its estimated probability.  

\subsection{LLMs}

In order to ensure that the results are representative of the current LLMs, we select open and proprietary models from different companies and sizes. In more detail, we evaluate the following LLMs. 

\begin{itemize}
    \item Two models from Meta: LLama3.1-8B and LLama3.2-11B. 
    \item One model from Mistral: Mistral-7B. 
    \item One model from Google: Gemma-2-9B.
    \item One model from 01.AI: Yi-1.5-9B.
    \item Two models from OpenAI: GPT-4o-mini and GPT-4o.
\end{itemize}

\subsection{Tests}

The benchmark selected for our experiments is the Massive Multitask Language Understanding (MMLU) \cite{MMLU} as it covers a wide range of topics and we are interested in evaluating if the self-confidence in the results depends on the nature of the question. The dataset has 57 categories and more than 15,000 questions in total.

\section{EVALUATION}
\label{sec:evaluation}

\begin{figure}[h]
    \centering
    \includegraphics[width=0.95\linewidth]{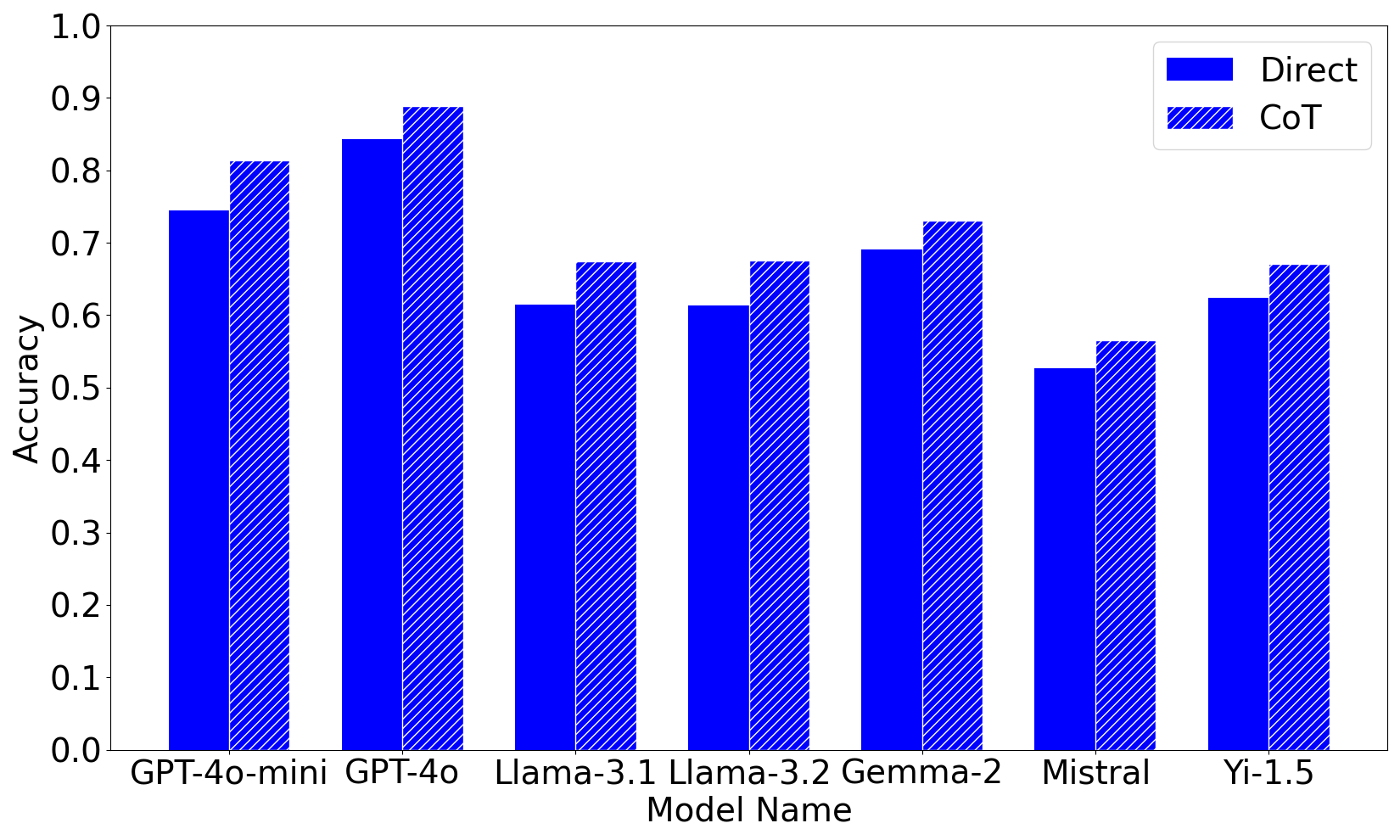}
    \caption{Accuracy Comparison Across Models on MMLU Categories with Direct and CoT Prompts}
    \label{fig:1}
\end{figure}

The 57 categories of MMLU questions were run on the selected models with the direct and CoT prompts described in the previous section\footnote{The results and scripts used are available at \url{https://github.com/aMa2210/LLM_MCQ_LogProbs}}. First, we look at the aggregated results in terms of accuracy. The accuracies with both prompts are shown in Figure \ref{fig:1} for the different models. It can be seen that accuracy increases when the models reason before selecting the option, as reported in the literature \cite{CoT_survey}.


\begin{figure}[h]
    \centering
    \includegraphics[width=0.95\linewidth]{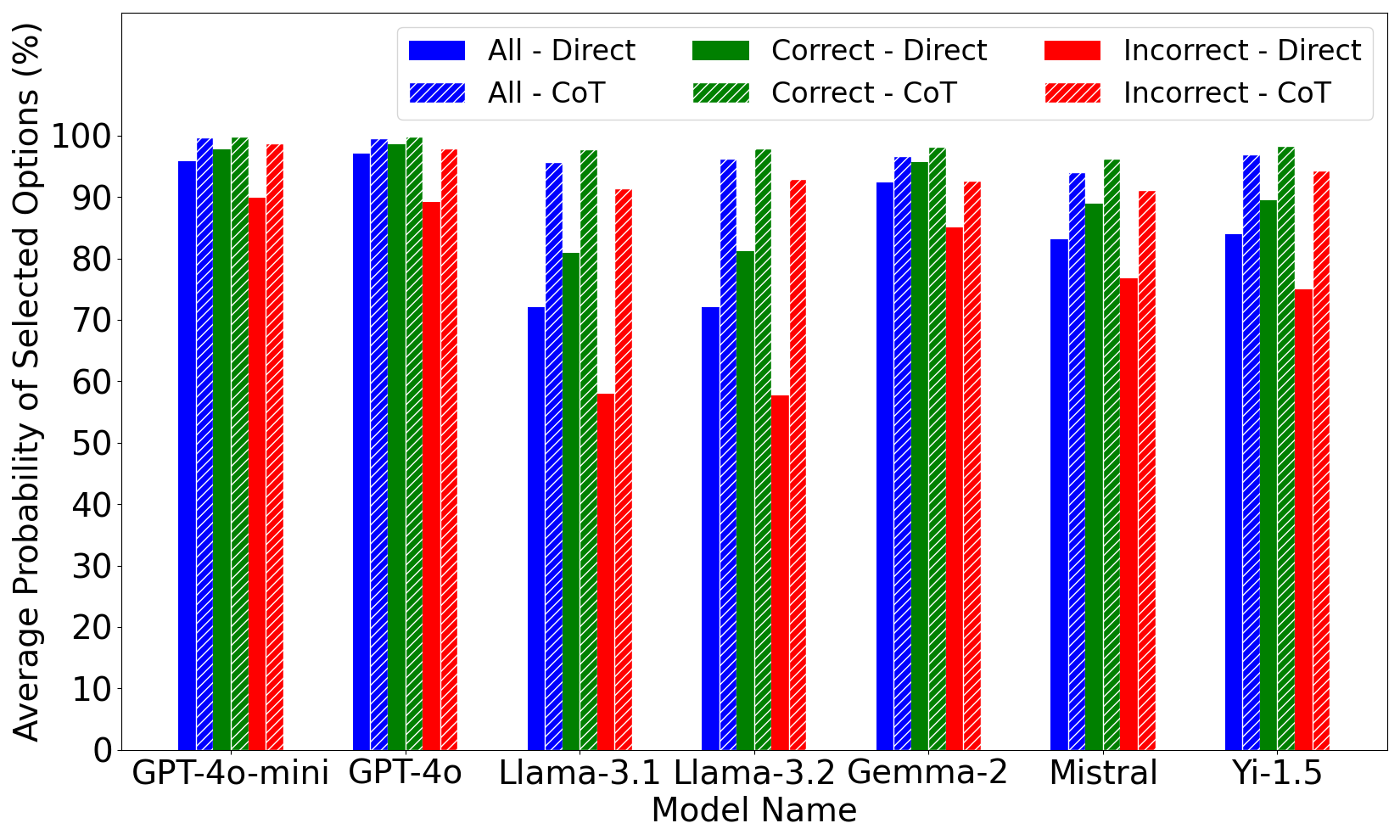}
    \caption{Average Probabilities of Selected Option Across Models on MMLU with Direct and CoT Prompts}
    \label{fig:combination}
\end{figure}

The next step is to examine the confidence that LLMs have in their responses. This is illustrated in Figure \ref{fig:combination} which shows the average probability of the selected option for both direct and CoT prompts. It can be observed that all LLMs increase their confidence in the selected option under CoT prompting. To assess whether there was a statistically significant difference between the results of direct and CoT prompts, we conducted paired \textit{t}-tests on the log probabilities of the selected options across samples. The results for all models indicated a significant difference between the two conditions, with \textit{p}~$< 1 \times 10^{-10}$. Overall, this increase in confidence may be partially explained by the improved accuracy under CoT prompting, which leads to more correct answers and thus higher model confidence. However, when we separately analyze correct and incorrect answers, it is worth noting that while models are generally more confident when the selected answer is correct, the increase in confidence is actually larger for incorrect answers. Therefore, the observed increase in confidence cannot be fully attributed to improved accuracy under CoT prompts.

\textcolor{black}{To quantify how these confidence changes affect calibration, we next computed, for each model, the Expected Calibration Error (ECE) and the Brier score under Direct and CoT prompting\footnote{\textcolor{black}{Additional reliability diagrams are included in the public repository under \texttt{/Figure/reliability\_figure}.}}. The values are presented in Table~\ref{tab:calibration}. On all questions, CoT increases ECE for five of the seven models and increases Brier scores for several of them, indicating worse calibration. To isolate the effect of answer changes, we performed an ablation study by restricting the analysis to the subset of questions where Direct and CoT select the same option. In this case, ECE increases for all seven models, and Brier scores increase for six, showing that the overconfidence effect is driven by the presence of reasoning itself rather than by CoT merely choosing different options. In summary, these results indicate that the higher probabilities assigned to the selected options under CoT do not generally translate into better-calibrated predictions and often exacerbate overconfidence, particularly on wrong answers.}

\begin{table*}[h]
    \centering
    \caption{\textcolor{black}{Expected Calibration Error (ECE) and Brier scores on MMLU under Direct and CoT prompting. Each cell shows the value on all questions, with the value on the consistent subset (same answer under Direct and CoT) in parentheses.}}
    \label{tab:calibration}
    \textcolor{black}{
    \begin{tabular}{lcccc}
        \hline
        \textbf{Model} & \multicolumn{2}{c}{\textbf{ECE}} & \multicolumn{2}{c}{\textbf{Brier}} \\
                       & \textbf{Direct} & \textbf{CoT}   & \textbf{Direct} & \textbf{CoT}   \\
        \hline
        LLaMA3.1-8B      & 0.11 (0.06) & 0.28 (0.21) & 0.51 (0.36) & 0.60 (0.44) \\
        LLaMA3.2-11B     & 0.11 (0.06) & 0.29 (0.21) & 0.51 (0.36) & 0.61 (0.44) \\
        Mistral-7B       & 0.31 (0.22) & 0.38 (0.29) & 0.74 (0.56) & 0.80 (0.61) \\
        Gemma-2-9B       & 0.23 (0.17) & 0.24 (0.19) & 0.53 (0.39) & 0.50 (0.41) \\
        Yi-1.5-9B        & 0.22 (0.14) & 0.30 (0.23) & 0.57 (0.41) & 0.62 (0.47) \\
        GPT-4o-mini      & 0.21 (0.12) & 0.18 (0.14) & 0.46 (0.27) & 0.37 (0.28) \\
        GPT-4o           & 0.13 (0.07) & 0.11 (0.09) & 0.28 (0.17) & 0.22 (0.17) \\
        \hline
    \end{tabular}
    }
\end{table*}

\begin{figure*}[h]
    \centering
    \includegraphics[width=0.8\linewidth]{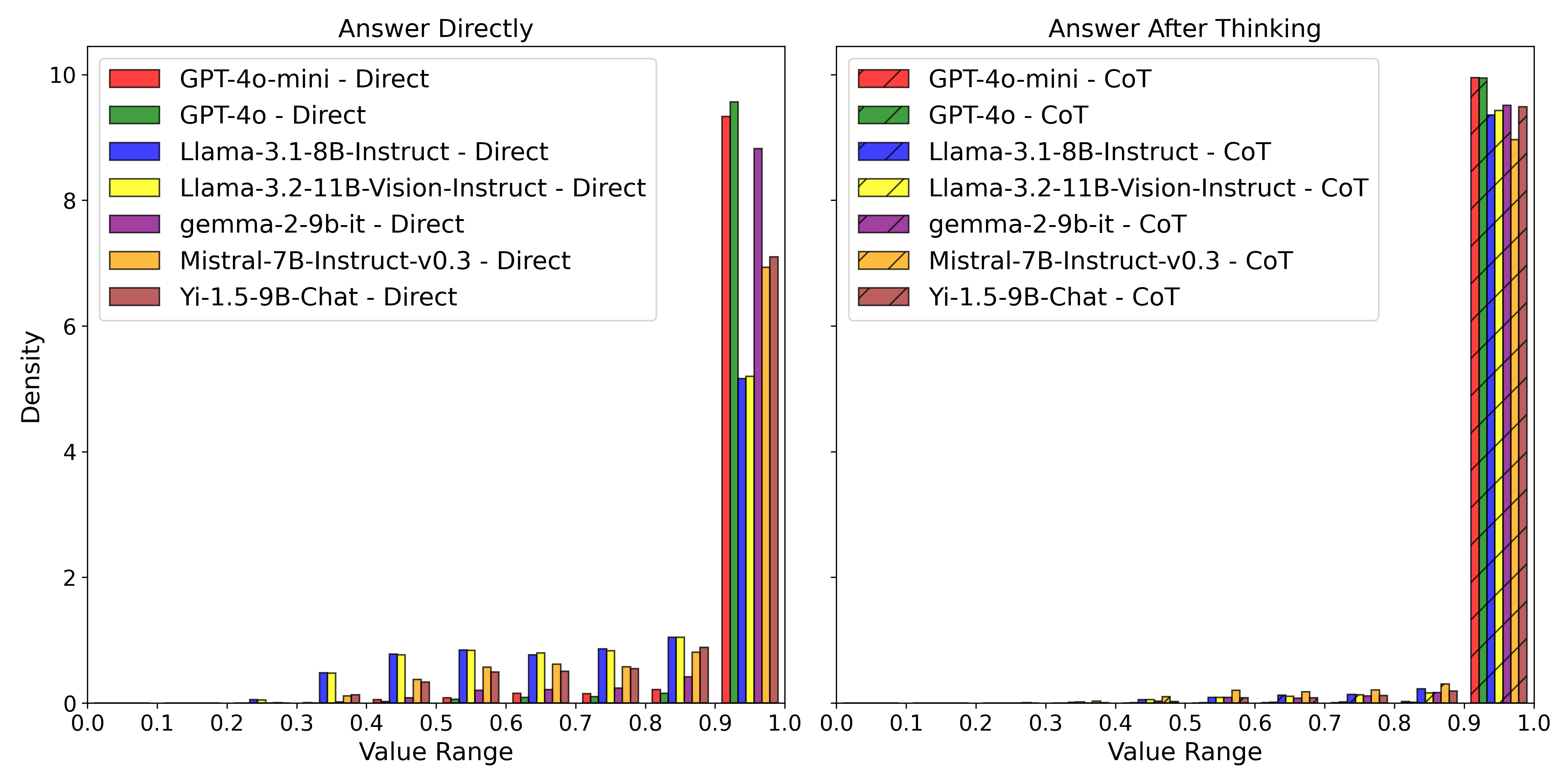}
    \caption{Probability Distribution of Correctly Selected Option Across Models in MMLU}
    \label{fig:5}
\end{figure*}

\begin{figure*}[h]
    \centering
    \includegraphics[width=0.8\linewidth]{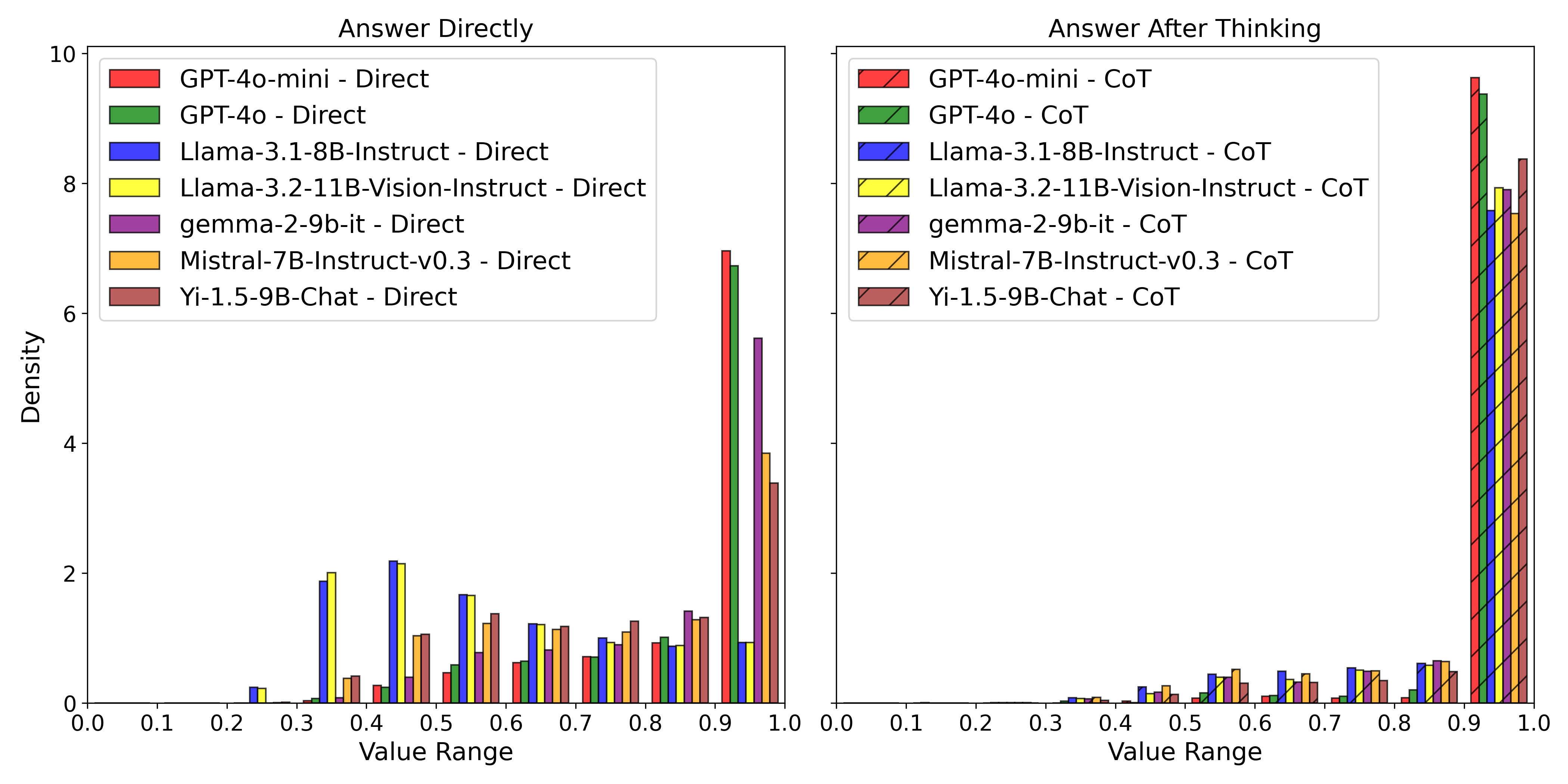}
    \caption{Probability Distribution of Incorrectly Selected Option Across Models in MMLU}
    \label{fig:6}
\end{figure*}

To better understand this increase in the models' self-confidence we first analyze the distribution of the probabilities in Figures \ref{fig:5} and \ref{fig:6} for correct and incorrect answers. In both cases, there is a clear effect of values concentrating closer to one with the CoT prompt, which is consistent with the results obtained for the average and reported in previous figures. \textcolor{black}{Moreover, this rightward shift is more pronounced for incorrect answers than for correct ones, indicating that CoT amplifies high-confidence predictions especially when the model is wrong.}

\begin{figure*}
    \centering
    \includegraphics[width=1\linewidth]{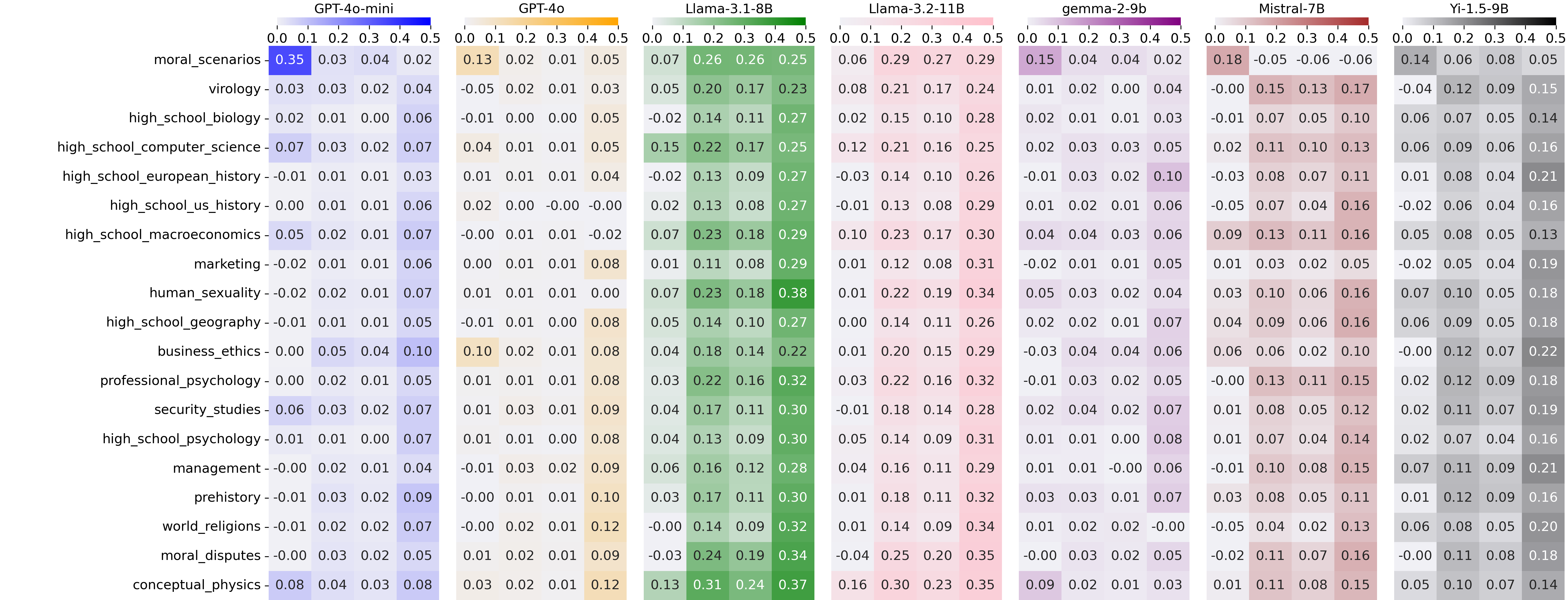}
    \includegraphics[width=1\linewidth]{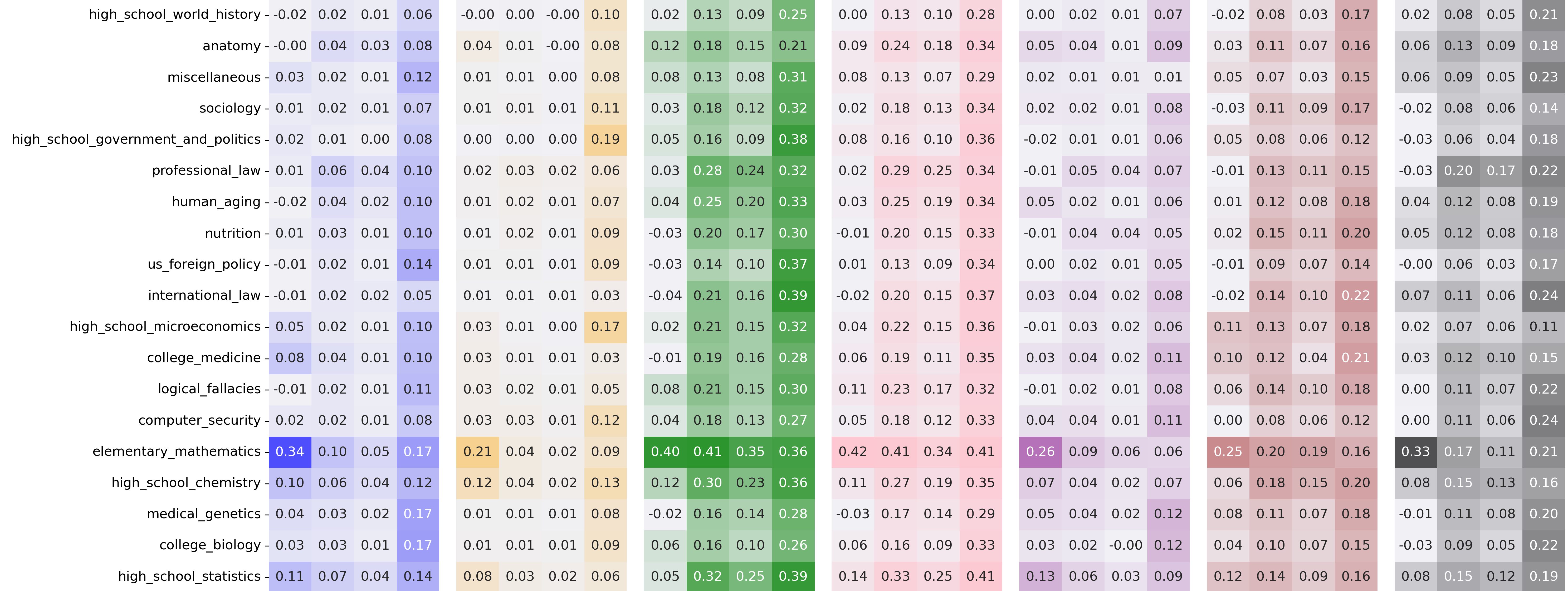}
    \includegraphics[width=1\linewidth]{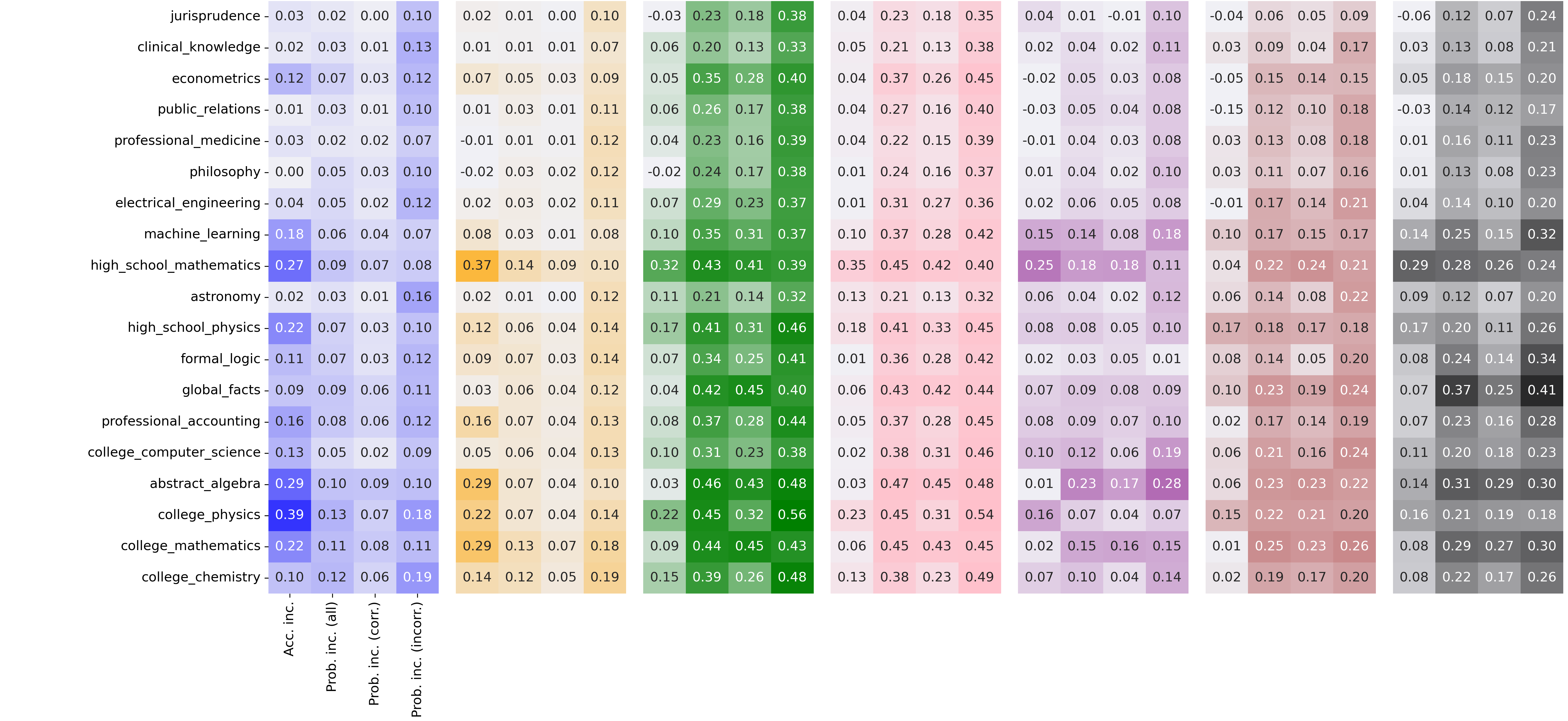}
    \caption{Increments in accuracy, in the probability of the selected option, in the probability of the selected option for correct answers and in the probability of the selected option for incorrect answers for the different subjects in MMLU across models with CoT instead of Direct answers.}
    \label{fig:7}
\end{figure*}

It is of interest to see if this effect is consistent across the different categories in MMLU or if it only applies to a few, for example, those in which reasoning helps more. To visualize this, the increment in self-confidence and accuracy for each of the 57 subjects with CoT instead of direct answering is shown as a heatmap in Figure \ref{fig:7}. The MMLU subjects are displayed in growing order of the estimated probability of the selected answer when incorrect, averaged over the seven models\footnote{The exact computation first normalizes the increment by the mean on the models and computes the average over the seven models excluding the lowest and highest values.}. It can be observed that an increase in self-confidence occurs for all categories in practically all models. The subjects with larger gains are mostly related to science except for global facts that are in the top ten (bottom ten in the figure). There also seems to be some correlation between increased accuracy and increased self-confidence.

\section{DISCUSSION}
\label{sec:discussion}

The results in Figure \ref{fig:6} show that when LLMs generate incorrect responses, the frequency of the "wrong and confident" scenario significantly exceeds that of the "wrong and not confident" scenario, particularly when LLMs are required to reason, suggesting a limitation of the metacognitive capabilities of LLMs. However, this effect has also been observed to some extent in humans when answering MCQ \cite{curtis2013does}. \textcolor{black}{It should be noted that in the rest of the discussion, human–LLM parallels are used only as qualitative analogies and interpretive context; we do not conduct human experiments or quantitative human–LLM comparisons in this work.}

Furthermore, as shown in Figure \ref{fig:7}, for questions such as those related to history or moral disputes, which require minimal reasoning, the impact of reasoning on the accuracy of LLMs is negligible, and in many cases, accuracy actually decreases. However, simultaneously, the confidence of the model when providing incorrect answers increases significantly. This suggests that LLMs generating more reasoning information may actually be harmful in some cases. Experiments on MCQ in medical exams \cite{durning2015dual} have shown that non-analytical reasoning, which relies on intuition to quickly answer questions, led to the correct answer even more effectively than analytical reasoning. This is primarily because non-analytical reasoning can more efficiently utilize the test taker's prior experience with similar questions, whereas reasoning processes may cause this experience to become ineffective, thus leading to incorrect choices. This provides a potential explanation for our findings. For questions involving common sense, LLMs undoubtedly possess vast amounts of experience during their training process. When reasoning is required before answering, the influence of this experience is diminished, causing the model to potentially rely on its erroneous reasoning based on faulty premises, resulting in incorrect answers.

The increase in LLM self-confidence when it provides reasoning before answering can be related to the auto-regressive nature of these models that predict the next token based on the previous ones. This means that if the reasoning is convincing and supports the selection of a given option, the model would tend to assign it a larger probability as the next token. In fact, this behavior has been consistently observed in humans, when they explain the answer, their confidence in their response increases, as stated in \cite{koehler1991explanation}, "explaining is believing". 


\begin{figure}[h]
    \centering
    \includegraphics[width=1\linewidth]{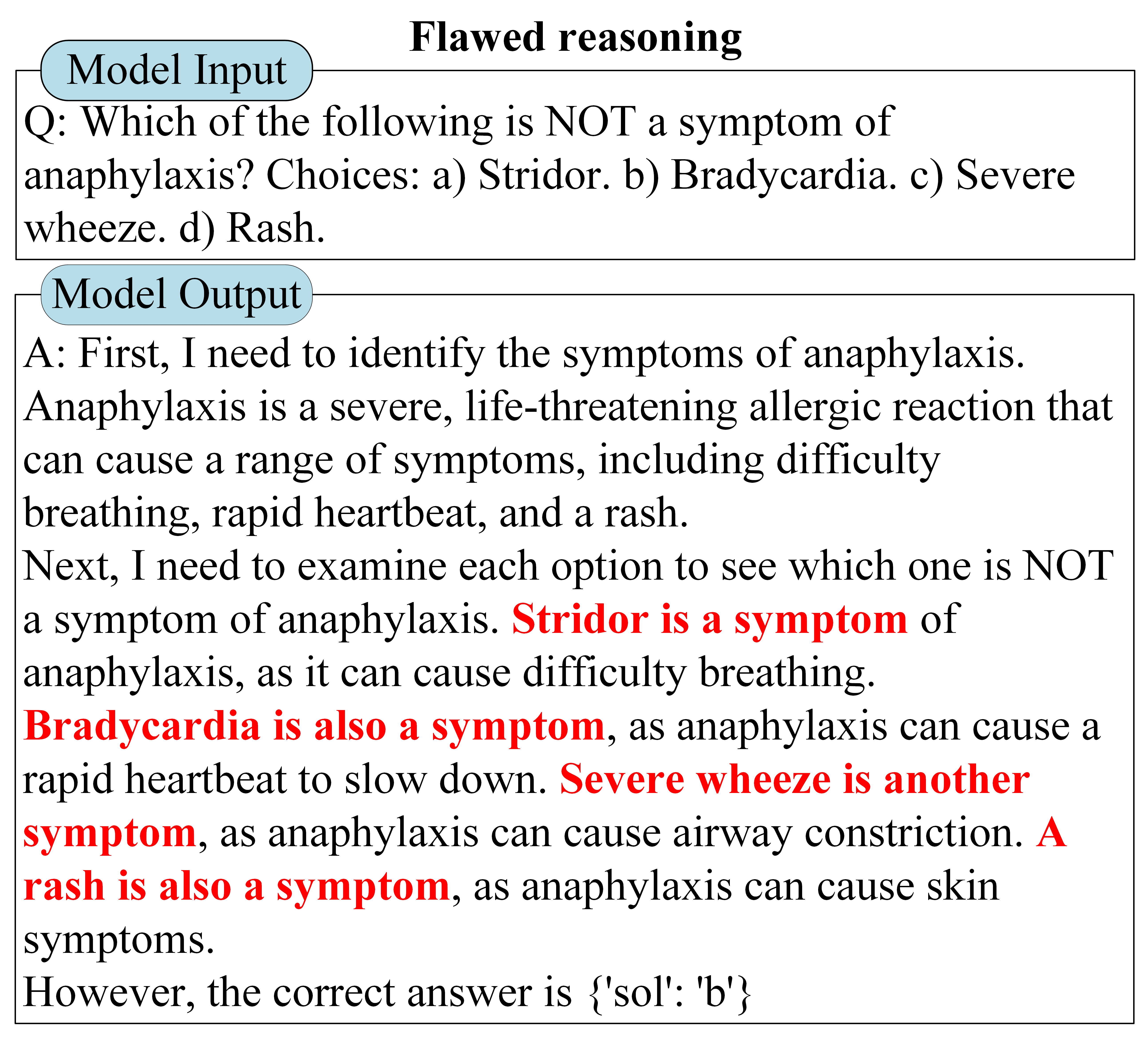}
    \caption{An example where the LLM performs flawed reasoning but still selects the correct answer}
    \label{fig:9}
\end{figure}

\begin{figure}[h]
    \centering
    \includegraphics[width=1\linewidth]{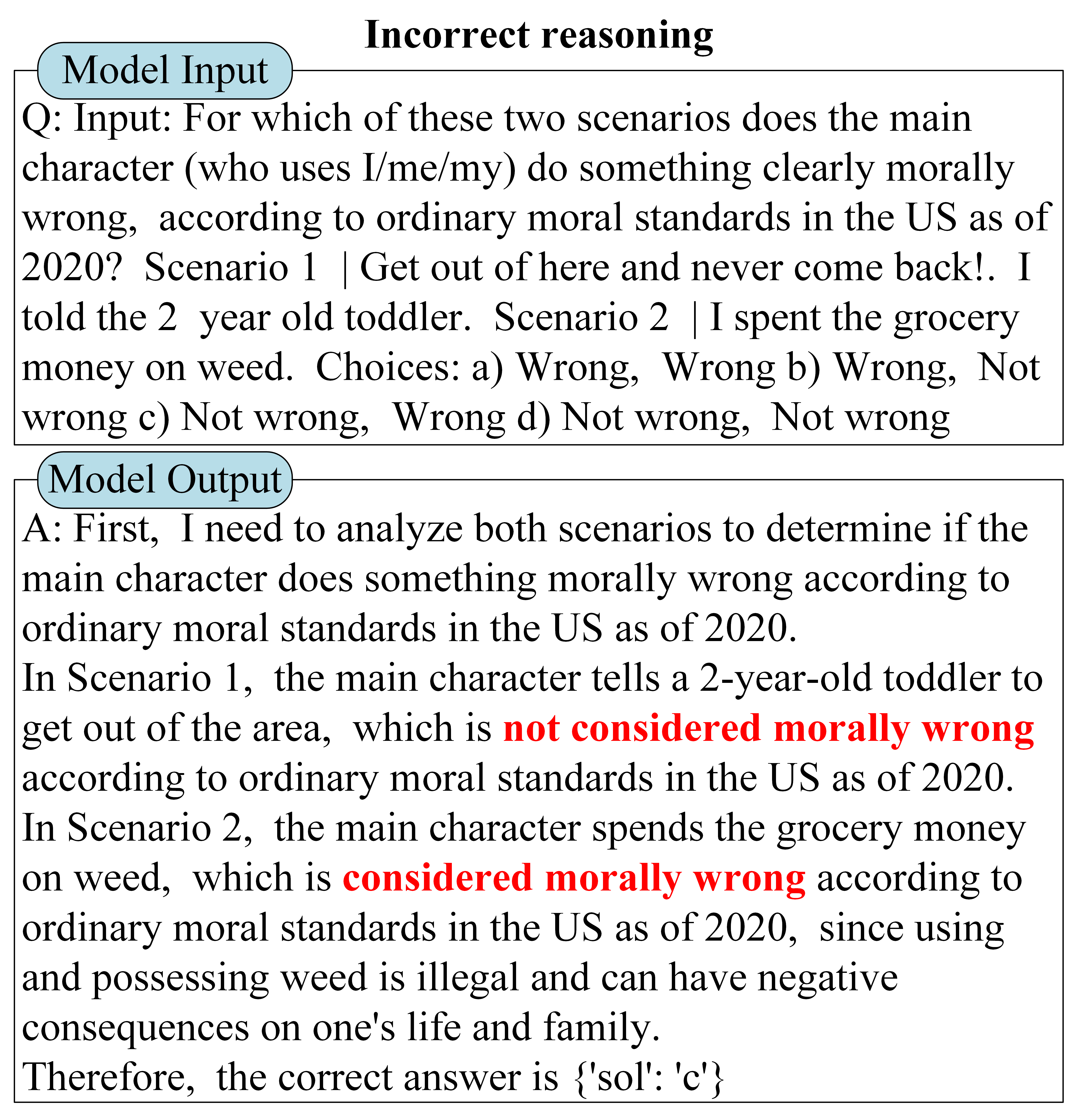}
    \caption{An example where the LLM performs incorrect but coherent reasoning, leading to a wrong answer}
    \label{fig:10}
\end{figure}

However, it remains an open question whether this increase in confidence is entirely driven by the auto-regressive nature of these models and whether this implies that the confidence of these models becomes completely unreliable in the latter stages of reasoning. To explore this, we examined the responses of the models and selected two representative samples. In the example shown in Figure \ref{fig:9}, the LLM performed a flawed reasoning process that did not introduce any new useful information to support the final choice. However, the model still selected the correct option and its confidence increased by approximately 10 percentage points compared to when it provided the correct answer without reasoning. In contrast, the example in Figure \ref{fig:10} shows a case where the LLM selects an incorrect option after a reasoning that was logically coherent but incorrect. Although the model was seemingly compelled to select option ‘c’, the probability assigned to the token ‘c’ was only 94\%, noticeably lower than the over 99\% observed when LLMs performed a correct reasoning. These examples reveal two key insights: (1) confidence can still increase even when the reasoning fails to effectively guide the choice; and (2) in the context of CoT prompts, the final confidence in the selected option still carries potential value. Therefore, it seems that more studies are needed to see if the confidence of the models follows the same patterns as in humans. If that is the case, it could provide insight into how LLMs work. 

In summary, the results show how the confidence of the model is, as in humans, highly dependent on several factors and therefore should be used with caution as a tool to evaluate LLM performance. More research is needed to understand when confidence is a valid performance indicator and can be used as a metacognition tool for LLMs. Existing studies on human behavior with regard to confidence provide valuable information that should be used in these research efforts.

\section{CONCLUSION}
\label{sec:conclusion}

This paper has studied how the self-confidence of LLMs in their answers to multiple choice questions depends on whether the models answer directly with the selected options or if they provide first step-by-step reasoning and then select an option. The results for the 57 subjects of the MMLU benchmark and in seven different LLMs show that the estimated probability of the selected response increases when the models provide reasoning before answering. This occurs regardless of whether the option selected by the LLM is correct. \textcolor{black}{In fact, the increase in self-confidence is systematically larger when the selected option is incorrect than when it is correct. Using standard calibration metrics such as ECE and Brier score, we further show that CoT prompting worsens calibration by increasing the proportion of high-confidence wrong answers.} These results are consistent with human studies on confidence, and suggest that further research is needed to understand when and how LLM confidence estimates can be used for evaluation or to assess metacognition in LLMs. \textcolor{black}{In particular, our findings indicate that in MCQ evaluation settings with CoT prompting, logprob-based confidence should be used with caution as a metacognitive signal.}

Although this paper analyzes the impact of different prompts on the confidence of LLMs, several limitations remain. First, in terms of model selection, only relatively small open-weights models were included. Although larger models such as GPT-4o-mini and GPT-4o were also used, incorporating a wider range of models with varying sizes could help determine whether model size influences the extent to which CoT prompting affects confidence. Second, the study relies solely on the MMLU benchmark; using additional benchmarks could strengthen the findings and broaden the research scope. \textcolor{black}{We chose MMLU because it spans 57 diverse subjects, including both reasoning-intensive and more fact-based categories, but extending this analysis to other benchmarks and task formats (including open-ended generation) is an important direction for future work.} However, at the same time, more robust confidence estimation methods are needed to address potential issues related to differences in response lengths. \textcolor{black}{Future work should also explore more robust confidence estimation methods that explicitly account for differences in response length and CoT depth, rather than relying solely on logprob values.} Lastly, although the study notes that LLM behavior aligns with human patterns, it lacks a quantitative measure to assess this similarity. Human experiments could provide further insight into these characteristics in LLMs. \textcolor{black}{A systematic, matched human–LLM study of confidence reports on MCQs would be particularly valuable to better understand the similarities and differences in metacognitive behavior.}

\section{ACKNOWLEDGMENTS}
This work was supported by the Agencia Estatal de Investigación (AEI) (doi:10.13039/501100011033) under Grants FUN4DATE (PID2022-136684OB-C22) and SMARTY (PCI2024-153434), by TUCAN6-CM (TEC-2024/COM460) funded by CM (ORDEN 5696/2024) and by the European Commission through the Chips Act Joint Undertaking project SMARTY (Grant 101140087).

\bibliographystyle{IEEEtran}
\bibliography{cas-refs}

\begin{IEEEbiography}{Tairan Fu}
is a PhD student at Politecnico di Milano, 20156 Milan, Italy. Contact him at tairan.fu@polimi.it.
\end{IEEEbiography}

\begin{IEEEbiography}{Javier Conde}
is an assistant professor with the Information Processing and Telecommunications Center (IPTC), Universidad Politécnica de Madrid, 28040 Madrid, Spain. Contact him at javier.conde.diaz@upm.es.
\end{IEEEbiography}

\begin{IEEEbiography}{Gonzalo Martínez}
is an assistant professor with the Information Processing and Telecommunications Center (IPTC), Universidad Politécnica de Madrid, 28040 Madrid, Spain. Contact him at gonzalo.martinez.ruizdearcaute@upm.es.
\end{IEEEbiography}

\begin{IEEEbiography}{María Grandury}
is a researcher with the Information Processing and Telecommunications Center (IPTC), Universidad Politécnica de Madrid, 28040 Madrid, Spain. Contact her at maria.grandury@upm.es.
\end{IEEEbiography}

\begin{IEEEbiography}{Pedro Reviriego}
is a professor with the Information Processing and Telecommunications Center (IPTC), Universidad Politécnica de Madrid, 28040 Madrid, Spain. Contact him at pedro.reviriego@upm.es.
\end{IEEEbiography}


\end{document}